\documentclass[11pt,a4paper]{article}
\usepackage[hyperref]{acl2017}
\usepackage{times}
\usepackage{latexsym}

\aclfinalcopy 

\usepackage{amsmath}
\usepackage{graphicx}
\usepackage{multirow}
\usepackage{algorithm}
\usepackage[noend]{algpseudocode}
\usepackage{relsize}
\usepackage{cleveref}
\usepackage[caption=false,font=footnotesize]{subfig}

\usepackage{url}

\newcommand{\ignore}[1]{}

\title{Learning how to learn: an adaptive dialogue agent for incrementally  learning visually grounded word meanings}

\author{Yanchao Yu\\ 
  Interaction Lab   \\
 Heriot-Watt University  \\
  {\tt y.yu@hw.ac.uk} \\ \And
  Arash Eshghi \\
 Interaction Lab   \\
 Heriot-Watt University  \\
  {\tt a.eshghi@hw.ac.uk} \\ \And
  Oliver Lemon \\
 Interaction Lab   \\
 Heriot-Watt University  \\
  {\tt o.lemon@hw.ac.uk}}
\date{}

\hypersetup{draft}

\begin{document}
\maketitle
\begin{abstract}

We present an optimised multi-modal dialogue agent for interactive learning of visually grounded word meanings from a human tutor, trained on real human-human tutoring data. Within a life-long interactive learning period, the agent, trained using Reinforcement Learning (RL), must be able to handle natural conversations with human users, and achieve  good learning performance (i.e.\ accuracy) while minimising human effort in the learning process. We train and evaluate this  system in interaction with a simulated human tutor, which is built on the BURCHAK corpus -- a Human-Human Dialogue dataset for the visual learning task. The results show that: 1) The learned policy can coherently interact with the simulated user to achieve the goal of the task (i.e.\ learning visual attributes of  objects, e.g.\ colour and shape); and 2) it finds a better trade-off between  classifier accuracy and tutoring costs than hand-crafted rule-based policies, including ones with dynamic policies. 

\end{abstract}

\section{Introduction}
\begin{figure}[!ht]
\centering

\resizebox{\linewidth}{!}{%
\begin{tabular}{|c|l|}\hline
\textbf{Image} & \multicolumn{1}{c|}{\textbf{Human-Human Dialogue}}\\
\hline

\raisebox{-0.5cm}{\includegraphics[width=1.8cm]{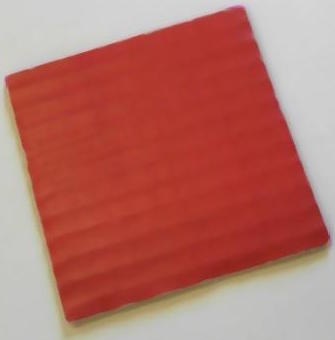}} & \raisebox{0.2cm}{\begin{tabular}{l}
T(utor): do you know this object?\\
L(earner): a suzuli ... wait no ... sako wakaki?\\
T: the color is right, but the shape is not.\\
L: oh, okay, so? \\
T: a burchak, burchak, sako burchak. \\
L: cool, got it.
\end{tabular}} \\

\hline

\raisebox{-0.5cm}{\includegraphics[width=1.8cm]{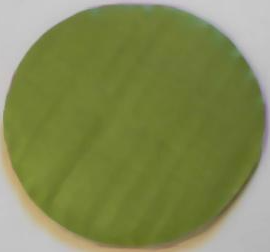}} & \raisebox{0.2cm}{\begin{tabular}{l}
L: what is this?\\
T: en ... a aylana suzili.\\
L: is aylana for color?\\
T: no, it's a shape.\\
L: so it is an suzili aylana, right?\\
T: yes. 
\end{tabular}}
\\ \hline
\end{tabular}
}
\vspace{-0.15cm}
\caption{Human-Human Example Dialogues in the BURCHAK Corpus \cite{yu-eshghi-lemon:2017:vl} \\ 
\begin{footnotesize}
\centering
  \noindent (`sako' for `red', `burchak' for `square', `suzuli' for `green', `aylana' for `circle', `wakaki' for `triangle')  
\end{footnotesize} \label{fig:human-human-example}}
\end{figure}
\vspace{-0.15cm}

As intelligent systems/robots are brought out of the laboratory and into the physical world, they must become capable of natural everyday conversation with their human users about their physical surroundings. Among other competencies, this involves the ability to learn and adapt mappings between words, phrases, and sentences in Natural Language (NL) and perceptual aspects of the external environment -- this is widely known as \textit{the grounding problem}. 

The grounding problem can be categorised into two distinct, but interdependent types of problem: 1) agent as a second-language learner: the agent needs to learn to ground (map) NL symbols onto their existing perceptual and lexical knowledge (e.g.\ a dictionary of pre-trained classifiers) as in e.g. \newcite{silberer-lapata:2014:P14-1,thomason-etal:2016:ijcai,kollar-etal:2013:robotics,matuszek-etal:2014:aaai}; and 2) the agent as a child: without any prior knowledge of perceptual categories, the agent must learn both the perceptual categories themselves and also how NL expressions map to these \citep{skocaj-etal:2016:tai,yu-eshghi-lemon:2016:sigdial}. Here, we concentrate on the latter scenario, where a system learns to identify and describe visual attributes (colour and shape in this case) through interaction with human tutors, incrementally, over time.

Previous work has approached the grounding problem using 
a variety of resources and approaches, for instance, either using annotated visual datasets \citep{silberer-lapata:2014:P14-1,socher-etal:2014:tacl,naim-etal:2015:naacl,alomari-etal:2016:kr,tellex-etal:2014:ml,matuszek-etal:2012:icml,matuszek-etal:2014:aaai}, or through interactions with other agents or real humans \cite{kollar-etal:2013:robotics,tellex-etal:2013:robotics,thomason-etal:2015:ijcai,thomason-etal:2016:ijcai,skocaj-etal:2016:tai,yu-eshghi-lemon:2016:sigdial}, where feedback from other agents is used to learn new concepts.

However, most of these systems, which ground NL symbols through interaction have two common, important drawbacks: 1) in order to achieve better performance (i.e.\ high accuracy), these systems require a high level of human involvement -- they always request feedback from human users, which 
might affect the quality of human answers and decrease the overall user experience in a life-long learning task; 2) Most of these approaches are not built/trained based on real human-human conversations, and therefore can't handle them. Natural human dialogue is generally more messy than either machine-machine or human-machine dialogue, containing natural dialogue phenomena that are notoriously difficult to capture, e.g.\  
\textit{self- corrections, repetitions and restarts, pauses, fillers, interruptions, and continuations} \cite{Purver.etal09a,Hough15}. 
Furthermore, they often exhibit much more variation than in their synthetic counterparts (see dialogue examples in Fig.\ \ref{fig:human-human-example}). 

In order to cope with the first problem, recent prior work \cite{yu-eshghi-lemon:2016:vl,yu-eshghi-lemon:2016:sigdial} has built multi-modal dialogue systems to investigate the effects of different dialogue strategies and capabilities on the overall learning performance. Their results have shown that, in order to achieve a good trade-off between learning performance and   human involvement, the agent must be able to take initiative in dialogues, take into account uncertainty of its predictions, as well as cope with natural human  conversation in the learning process. However, their systems are built based on hand-crafted, synthetic dialogue examples rather than real human-human dialogues.

In this paper, we extend this work to introduce an adaptive visual-attribute learning agent trained using Reinforcement Learning (RL). The agent, trained with a multi-objective policy, is capable not only of properly learning novel visual objects/attributes through interaction with human tutors, but also of efficiently minimising human involvement in the learning process. It can achieve equivalent/comparable learning performance (i.e.\ accuracy) to a fully-supervised system,  but with less tutoring effort. The dialogue control policy is trained on the BURCHAK Human-Human Dialogue dataset \cite{yu-eshghi-lemon:2017:vl}, consisting of conversations between a human `tutor' and a human `learner' on a visual attribute learning task. The dataset includes a wide range of natural, {\it incremental} dialogue phenomena (such as overlapping turns, self-correction, repetition, fillers, and continuations), as well as considerable variation in the dialogue strategies used by the tutors and the learners.

Here we compare the new optimised learning agent to  rule-based agents with and without  adaptive confidence thresholds (see section \ref{adaptconf}).
 The results show that the RL-based learning agent outperforms the rule-based systems by finding a better trade-off between learning performance and the tutoring effort/cost.  

\section{Related Work \label{sec:literature}}
In this section, we review some of the work that has addressed the language grounding problem generally. The problem of grounding NL in perception has received very considerable attention in the computational literature recently. On the one hand, there is work that only addresses the grounding problem implicitly/indirectly: in this category of work is the large literature on image and video captioning systems that learn to associate an image or video with NL descriptions \cite{silberer-lapata:2014:P14-1,bruni2014multimodal,socher-etal:2014:tacl,naim-etal:2015:naacl,alomari-etal:2016:kr}. This line of work uses various forms of neural modeling to discover the association between information from multiple modalities. This often works by projecting vector representations from the different modalities (e.g.\ vision and language) into the  same space in order to retrieve one from  the other.  Importantly, these models are holistic in that they learn to use NL symbols in specific tasks without any explicit encoding of the symbol-perception link, so that this relationship remains implicit and indirect.

On the other hand, other models assume a much more explicit connection between symbols (either words or predicate symbols of some logical language) and perceptions \cite{kennington-schlangen:2015:acl,yu-eshghi-lemon:2016:sigdial,skocaj-etal:2016:tai,Dobnik.etal14,matuszek-etal:2014:aaai}. In this line of work, representations are both compositional and transparent, with their constituent atomic parts grounded individually in perceptual classifiers. Our work in this paper is in the spirit of the latter.

\vspace{-0.05cm}
Another dimension along which work on grounding can be compared is whether groundings are learned offline (e.g. from images or videos annotated with descriptions or definite reference expressions as in \cite{kennington-schlangen:2015:acl,socher-etal:2014:tacl}) or from live interaction as in, e.g. \cite{skocaj-etal:2016:tai,yu-eshghi-lemon:2015:vl,yu-eshghi-lemon:2016:sigdial,das-etal:2017:corr,das-etal:2016:corr,vries-etal:2016:corr,thomason-etal:2015:ijcai,thomason-etal:2016:ijcai,tellex-etal:2013:robotics}. The latter, which we do here, is clearly more appropriate for multimodal systems or robots that are expected to continuously, and incrementally learn from the environment and their users.

\vspace{-0.05cm}
Multi-modal, interactive systems that involve grounded language are either: (1) \textit{rule-based} as in e.g. \newcite{skocaj-etal:2016:tai,yu-eshghi-lemon:2016:vl,thomason-etal:2015:ijcai,thomason-etal:2016:ijcai,tellex-etal:2013:robotics,schlangen:2016:semdial}: in such systems, the dialogue control policy is hand-crafted, and therefore these systems are \textit{static}, cannot adapt, and are less robust; or (2) \textit{optimised} as in e.g.\  \newcite{yu-eshghi-lemon:2016:sigdial,mohan-etal:2012:aaai,whitney-etal:2013:icra,das-etal:2017:corr}: in contrast such systems are learned from data, and live interaction with their users; they can thus \textit{adapt} their behaviour dynamically not only to particular dialogue histories, but also to the specific information they have in another modality (e.g.\ a particular image or video). 



\vspace{-0.05cm}
Ideally, such interactive systems ought to be able to handle natural, spontaneous human dialogue. 
However, most work on interactive language grounding learn their systems from synthetic, hand-made dialogues or simulations which lack both in variation and the kinds of dialogue phenomena that occur in everyday conversation; they thus lead to systems which are not robust and cannot handle everyday conversation \cite{yu-eshghi-lemon:2016:sigdial,skocaj-etal:2016:tai,Yu.etal16semdial}. In this paper, we try to change this by training an adaptive learning agent from \textit{human-human dialogues in a visual  attribute learning task}.

\vspace{-0.05cm}
Given the above, what we achieve here is: we have trained an adaptive attribute-learning dialogue policy from realistic human-human conversations that learns to optimise the trade-off between a learning/grounding performance (\textit{Accuracy}) and costs form human tutors,in effect doing a form of active learning.    



\vspace{-0.2cm}
\section{Learning How to Learn Visual Attributes: an Adaptive Dialogue Agent}\label{sec:agent}
We build a multimodal and teachable system that supports a visual attribute (e.g.\ colour and shape) learning process through natural conversational interaction with human tutors (see Fig.\ \ref{fig:human-human-example} for example dialogues), where the tutor and the learner interactively exchange information about the visual attributes of an object they can both see. Here we use Reinforcement Learning for policy optimisation for the learner side (see below Section \ref{sec:learn-agent}). The tutor side is simulated in a data-driven fashion using human-human dialogue data (see below, Sections ~\ref{sec:corpus} \& ~\ref{sec:sim}).


\vspace{-0.1cm}
\subsection{Overall System Architecture}\label{sec:arch}
The system architecture loosely follows that of \newcite{yu-eshghi-lemon:2016:sigdial}, and employs two core modules: 

\vspace{-0.1cm}
\paragraph{Vision Module}   produces visual attribute predictions, using two base feature categories, i.e.\ the HSV colour space for colour attributes, and a `bag of visual words' (i.e.\ PHOW descriptors) for the object shapes/class.  It consists of  a set of binary classifiers - Logistic Regression SVM classifiers with Stochastic Gradient Descent (SGD) \cite{zhang2004solving} -- to incrementally learn attribute predictions. The visual classifiers ground visual attribute words such as `red', `circle' etc.\ that appear as parameters of the Dialogue Acts used in the system. 

\vspace{-0.1cm}
\paragraph{Dialogue Module} that
implements a dialogue system with a classical architecture, composed of Dialogue Management (DM), Natural Language Understanding (NLU) and Generation (NLG) components. The components interact via Dialogue Act representations (e.g.\ \texttt{inform(color=red),}\texttt{ask(shape)}). It is these action representations that are grounded in the visual classifiers that reside in the vision module.
The DM relies on an adaptive policy that is learned using RL. The policy is trained to: 1) handle natural interactions with humans and to produce coherent dialogues; and 2) optimise the trade-off between accuracy of visual classifiers and the cost of the dialogue to the tutor.



\vspace{-0.1cm}
\subsection{Adaptive Learning Agent with Hierarchical MDP \label{sec:learn-agent}}
Given the visual attribute learning task, the smart agent must learn novel visual objects/attributes as accurately as possible through natural interactions with real humans, but meanwhile it should attempt to minimise the human involvement as much as possible in this life-long learning process. We formulate this interactive learning task into two sub-tasks, which are trained using Reinforcement Learning with a hierarchical Markov Decision Process (MDP), consisting of two interdependent MDPs (sections \ref{adaptconf} and \ref{naturalint}): 

\subsubsection{Adaptive Confidence Threshold}\label{adaptconf}
Following previous work \citep{yu-eshghi-lemon:2016:sigdial}, we also here use a positive confidence threshold: this is a threshold which determines when the agent believes its own predictions. This threshold plays an essential role in achieving the trade-off between the learning performance and the tutoring cost, since the agent's behaviour, e.g. whether to seek feedback from the tutor, is dependent on this threshold. A form of \textit{active learning} is taking place: the learner only asks a question about an attribute if it isn't confident enough already about that attribute.

Here, we learn an adaptive strategy that aims at maximising the overall learning performance simultaneously, by properly adjusting the positive confidence threshold in the range of 0.65 to 0.95. We train the optimization using a RL library -- Burlap 
\citep{burlap_james} as follows, in detail:

\vspace{-0.1cm}
\paragraph{State Space} The adaptive-threshold MDP initialises a 3-dimensional state space defined by $Num_{Instance}$, $Threshold_{cur}$, and $deltaAcc$, where $Num_{Instance}$ represents how many visual objects/images have been seen (the number of instances will be clustered into 50 bins, each bin contains 10 visual instances); $Threshold_{cur}$ represents the positive threshold the agent is currently applying; and $deltaAcc$ represents, after seeing each 10 instances, whether the classifier accuracy increases, decreases or keep constant comparing to the previous bin. The $deltaAcc$ is configured into three levels, (see Eq.\ref{equ:deltaAcc})

\vspace{-0.1cm}
\begin{equation}\label{equ:deltaAcc}
 	deltaAcc =
    \begin{cases}
      1, & \text{if}\ \Delta Acc > 0   \\
      0, & \text{else if}\ \Delta Acc = 0 \\
      -1, & \text{otherwise}
    \end{cases}
\end{equation}

\vspace{-0.1cm}
\paragraph{Action Selection}  the actions were either to increase or decrease the confidence threshold by 0.05, or keep it the same.

\vspace{-0.1cm}
\paragraph{Reward signal} The reward function for the learning tasks is given by a local function $R_{local}$. This local reward signal was directly proportional to the agent’s delta accuracy over the previous Learning Step (10 training instances, see above). The single training episode will be terminated once the agent goes through 500 instances. 

\begin{table*}[!ht]
\centering
\begin{tabular}{lcl}
\hline
\textbf{Dialogue Capability} & \textbf{Speaker} & \multicolumn{1}{c}{\textbf{Annotation Tag}} \\
\hline \hline
Listen & Tutor/Learner & Listen() \\
Inform & Tutor/Leaner & Inform(colour:sako\&shape:burchak) \\
Question\_asking & Tutor/Leaner & Ask(colour), Ask(shape), Ask(colour\&shape) \\
Question-answering & Tutor/Leaner & Inform(colour:sako), Polar(shape:burchak) \\
Acknowledgement & Tutor/Learner & Ack(), Ack(colour)\\
Rejection & Tutor & Reject(), Reject(shape) \\
Focus & Tutor & Focus(colour), Focus(shape) \\
Clarification & Tutor & CLr() \\
Clarification-request & Learner & CLrRequest() \\
Help-offer & Tutor & Help() \\
Help-request & Learner & HelpRequest() \\
Checking & Tutor & Check() \\
Repetition-request & Tutor & Repeat() \\
Retry-request & Tutor & Retry() \\
\hline
\end{tabular}%
\caption{List of Dialogue Capabilities/Actions and Corresponding Annotations in the Corpus}\label{tab:annotateCapability}
\end{table*}
\vspace{-0.2cm}

\subsubsection{Natural Interaction}\label{naturalint}

The second sub-task aims at learning an optimised dialogue strategy that allows the system to achieve the learning task (i.e.\ learn new visual attributes) through natural, human-like conversations. 

\vspace{-0.1cm}
\paragraph{State Space} The dialogue agent  initialises a 4-dimensional state space defined by ($C_{state}$, $S_{state}$, $preDAts$, $preContext$), where $C_{state}$ and $S_{state}$ are the status of visual predictions for the colour and shape attributes respectively (where the status is determined by the prediction score ($conf.$) and the adaptive confidence threshold ($posThd.$) described above (see Eq.\ref{equ:status})), the $preDAts$ represents the previous dialogue actions from the tutor response, and the $preContext$ represents which attribute categories (e.g.\ colour, shape or both) were talked about in the context history. 

\vspace{-0.1cm}
\begin{equation}\label{equ:status}
 	State =
    \begin{cases}
      2, & \text{if}\ conf. \geq posThd \\
      1, & \text{else if}\  0.5 < conf.< posThd. \\
      0, & \text{otherwise}
    \end{cases}
\end{equation}

\noindent i.e. $C_{state}$ or $S_{state}$ will be updated to 2 also when the related knowledge has been provided by the tutor.

\vspace{-0.1cm}
\paragraph{Action Selection} The actions were chosen based on the statistics of the dialog action frequency occurred from the BURCHAK corpus, including \textit{question-asking(for WH questions or polar questions)}, \textit{inform}, \textit{acknowledgment}, as well as \textit{listening}. These actions can be applied for either specific single attribute or both. The action of \textit{inform} can be separated into two sub-actions according to whether the prediction score is greater than 0.5 (i.e. \textit{polar question}) or not (i.e.\ \textit{doNotKnow}).

\vspace{-0.1cm}
\paragraph{Reward signal} The reward function for the learning tasks is given by a global function $R_{global}$ (see Eq.\ref{equ:reward}). The dialogue will be terminated when both colour and shape knowledge are either taught by human tutors or known with high confidence scores. 

\vspace{-0.1cm}
\begin{equation}\label{equ:reward}
 	R_{global} = 10 - C_{ost} - penal.;
\end{equation}

where $C_{ost}$ represents the cumulative cost by the tutor (see more details about this setup in Section \ref{sec:metric}) in a single dialogue, and $penal.$ penalizes all performed actions which cannot respond to the user properly. 

\vspace{0.4cm}
\noindent i.e. we applied the {\bf SARSA algorithm} \cite{Sutton.Barto98} for learning the multi-MDP learning agent with each episode defined as a complete dialogue for an object. It was configured with a $\xi-$Greedy exploration rate of 0.2 and a discount factor of 1. 


\section{Human-Human Dialogue Corpus: BURCHAK \label{sec:corpus}}

BURCHAK \cite{yu-eshghi-lemon:2017:vl} is a freely available Human-Human Dialogue dataset consisting of 177 dialogues between real human users on the task of interactively learning visual attributes.

\vspace{-0.1cm}
\paragraph{The DiET experimental toolkit} These dialogue were collected using a new \emph{incremental variation} of the DiET chat-tool developed by \citep{Healey.etal03,Mills.HealeySubmitted}, which allows two or more participants to communicate in a shared chat window. It supports live, fine-grained and highly local experimental manipulations of ongoing human-human conversation (see e.g. \cite{Eshghi.Healey15}). The chat-tool is designed to support, elicit, and record at a fine-grained level, dialogues that resemble face-to-face dialogue in that turns are: (1) constructed and displayed incrementally as they are typed; (2) transient; (3) potentially overlapping; (4) not editable, i.e.\ deletion is not permitted.
 
\vspace{-0.1cm}
\paragraph{Task} The learning/tutoring task given to the participants involves a pair of participants who talk about visual attributes (e.g.\ colour and shape) through a series of visual objects. The overall goal of this task is for the learner to discover groundings between visual attribute words and aspects in the physical world through interaction. However, since humans have already known all groundings, such as ``red'' and ``square'', the task is assumed in a second-language learning scenario, where each visual attribute, instead of standard English words, is assigned to a new unknown word in a made-up language (see examples in Fig. \ref{fig:human-human-example}). (see more details in \cite{yu-eshghi-lemon:2017:vl})

\vspace{-0.1cm}
\paragraph{Dialogue Phenomena} As the chat-tool is designed to resemble face-to-face dialogue, the most important challenge of this BURCHAK is that it refers to a wide range of natural, incremental dialogue phenomena, such as overlapping, self-correction and repetition, filler as well as continuation (Fig. \ref{fig:human-human-example}). On the other hand, BURCHAK, which focuses on the visual attribute learning task, offers a list of interesting task-oriented dialogue strategies (e.g. initiative, context-dependency and knowledge-acquisition) and capabilities, such as inform, question-asking and answering, listen (no act), as well as acknowledgement and rejection. Each dialogue action contains a huge variations in the realistic conversation. All dialogue actions are tagged in the dataset (as shown in Table \ref{tab:annotateCapability}).

\noindent i.e.\ we have trained and evaluated the optimised learning agents on the cleaned-up version of this corpus, in which spelling mistakes, emoticons, as well as some snippets of conversations where the participant misunderstood the task have been corrected or removed.   

\section{Experiment Setup \label{sec:expt}}

In this section, we follow previous work \cite{yu-eshghi-lemon:2016:sigdial} to compare the trained RL-based learning agent with a rule-based system with the best performance (i.e.\ an agent which takes the initiative in dialogues, takes into account its changing confidence about its predictions, and is also  able to process natural, human-like dialogues) from previous work. Instead of using hand-crafted dialogue examples as before, both the RL-based system and the rule-based system are trained/developed against a simulated user, itself trained from the BURCHAK dialogue data set as above. For learning simple visual attributes (e.g. \textit{``red''} and \textit{``square''}), we use the same hand-made visual object dataset from \newcite{yu-eshghi-lemon:2016:sigdial}.

In order to further investigate the effects of the optimised adaptive confidence threshold on the learning performance, we build the rule-based system under three different settings, i.e.\ with a constant threshold ($0.95$) (see \textit{blue} curve in Fig. \ref{fig:results}), with a hand-crafted adaptive threshold which drops by 0.05 after each 10 instances (\textit{grey} curve in Fig. \ref{fig:results}), and with a hand-crafted adaptive threshold which drops by 0.01 after each 10 instances (\textit{orange} curve in Fig. \ref{fig:results}). 

\subsection{Evaluation Metrics \label{sec:metric}}
To compare the optimised and the rule-based learning agents, and also further investigate how the adaptive threshold affect the learning process, we follows the evaluate metrics from the previous work (see \cite{yu-eshghi-lemon:2016:sigdial}) considering both the cost to the tutor and the accuracy of the learned meanings, i.e.\ the classifiers that ground our colour and shape concepts.

\vspace{-0.1cm}
\paragraph{Cost} The cost measure reflects the effort needed by a human tutor in interacting with the system. Skocaj et.\ al.\ \shortcite{Skocaj2009a} point out that a comprehensive teachable system should learn as autonomously as possible,  rather than involving the human tutor too frequently. There are several possible costs that the tutor might incur
: $C_{inf}$ refers to the cost $(i.e.\ 5 ~points)$ of the tutor providing information on a single attribute concept (e.g.\ ``this is red'' or ``this is a square"); $C_{ack}$ is the cost $(i.e.\ 0.5)$ for a simple confirmation (like ``yes", ``right") or rejection (such as ``no''); $C_{crt}$ is the cost of correction for a single concept (e.g.\ ``no, it is blue" or ``no, it is a circle"). We associate a higher cost $(i.e.\ 5)$ with correction of statements than that of polar questions. This is to penalise the learning agent when it confidently makes a false statement --  thereby incorporating an aspect of trust in the metric (humans will not trust systems which confidently make false statements). 

\noindent i.e. differently to the previous evaluation metrics, we do not take into account the costs of parsing and producing utterances 

\vspace{-0.1cm}
\paragraph{Learning Performance} As mentioned above, an efficient learner dialogue policy should consider both classification accuracy  and tutor effort (Cost). We thus define an integrated measure -- the \textit{Overall Performance Ratio} ($R_{perf}$) -- that we use to compare the learner's overall performance across the different conditions:

\vspace{-0.4cm}
\begin{displaymath}
 R_{perf} = \frac{\Delta Acc}{C_{tutor}}
\end{displaymath}

\noindent i.e.\ the increase in accuracy per unit of the cost, or equivalently the gradient of the curve in Fig.\ 2c.\ We seek dialogue strategies that maximise this.

\subsection{User Simulation}\label{sec:sim}

In order to train and evaluate these learning agents, we build an user simulation using a generic n-gram framework (see \cite{yu-eshghi-lemon:2017:vl}) on the BURCHAK corpus. This user framework takes as input the sequence of N most recent words in the dialogue,  as well as some optional additional conditions, and then outputs the next user response on multiple levels as required, e.g. full utterance, a sequence of dialogue actions, or even a sequence of single word outputs for incremental dialogue. Differently to other existing user simulations, this framework aims at not only resembling user strategies and capabilities in realistic conversations, but also at simulating incremental dialogue phenomena, e.g. self-repair and repetition, and pauses, as well as fillers. In this paper, we created an action-based user model that predict the next user response in a sequence of dialogue actions. The simulator then produces a full utterance by following the statistics of utterance templates for each predicted action. 


\ignore{
\subsection{Visual Object Dataset}

For comparing the learning performance (i.e.\ classifier accuracy) of different learning agents, we evaluate these agents on a collection of 600 images of simple hand-made objects\footnote{All data from  this paper will be made freely available.}. The goal of this system was to learn simple visual attributes (e.g.\ colour and shape) from these simple objects (see \cite{yu-eshghi-lemon:2016:vl}). There are nine attributes considered in this dataset: 6 colours (black, blue, green, orange, purple and red) and 3 shapes (circle, square and triangle). As background noise might interfere with the ability of object segmentation and extraction, we build images containing only one object within a white background. We keep a relative balance on the number of instances for each attribute in the dataset.  

}

\subsection{Results \label{sec:result}}
Table \ref{tab:example_dialogue} shows example interactions between the learned RL agent and the simulated tutor on the learning task. The dialogue agent learned to take the initiative and constantly produces coherent conversations through the learning process. 

\begin{figure*}[!ht]
   \subfloat[Accuracy\label{fig:acc_expt}]
  {\includegraphics[width=.5\linewidth]{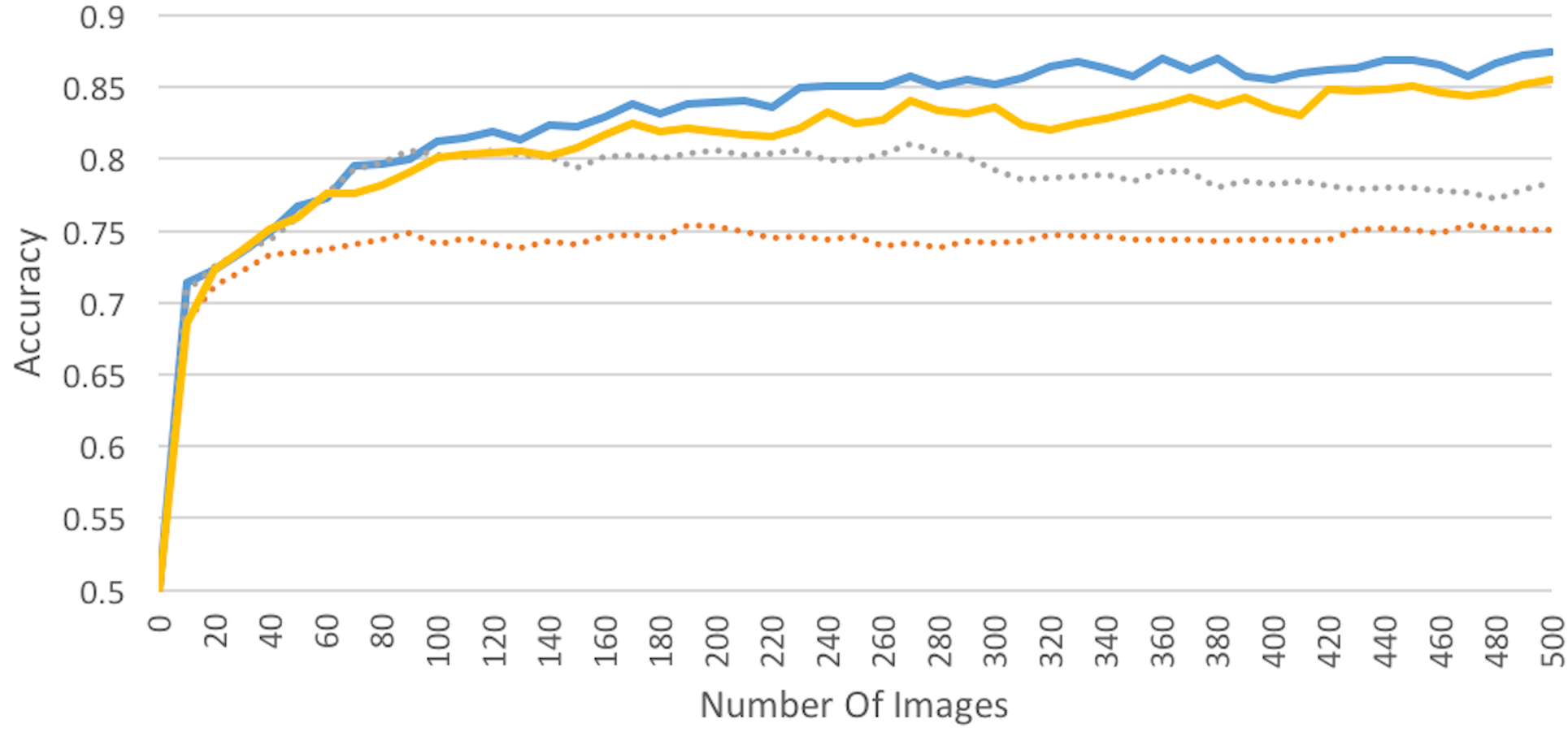}}\hfill
  \subfloat[Tutoring Cost\label{fig:cost_expt}]
  {\includegraphics[width=.5\linewidth]{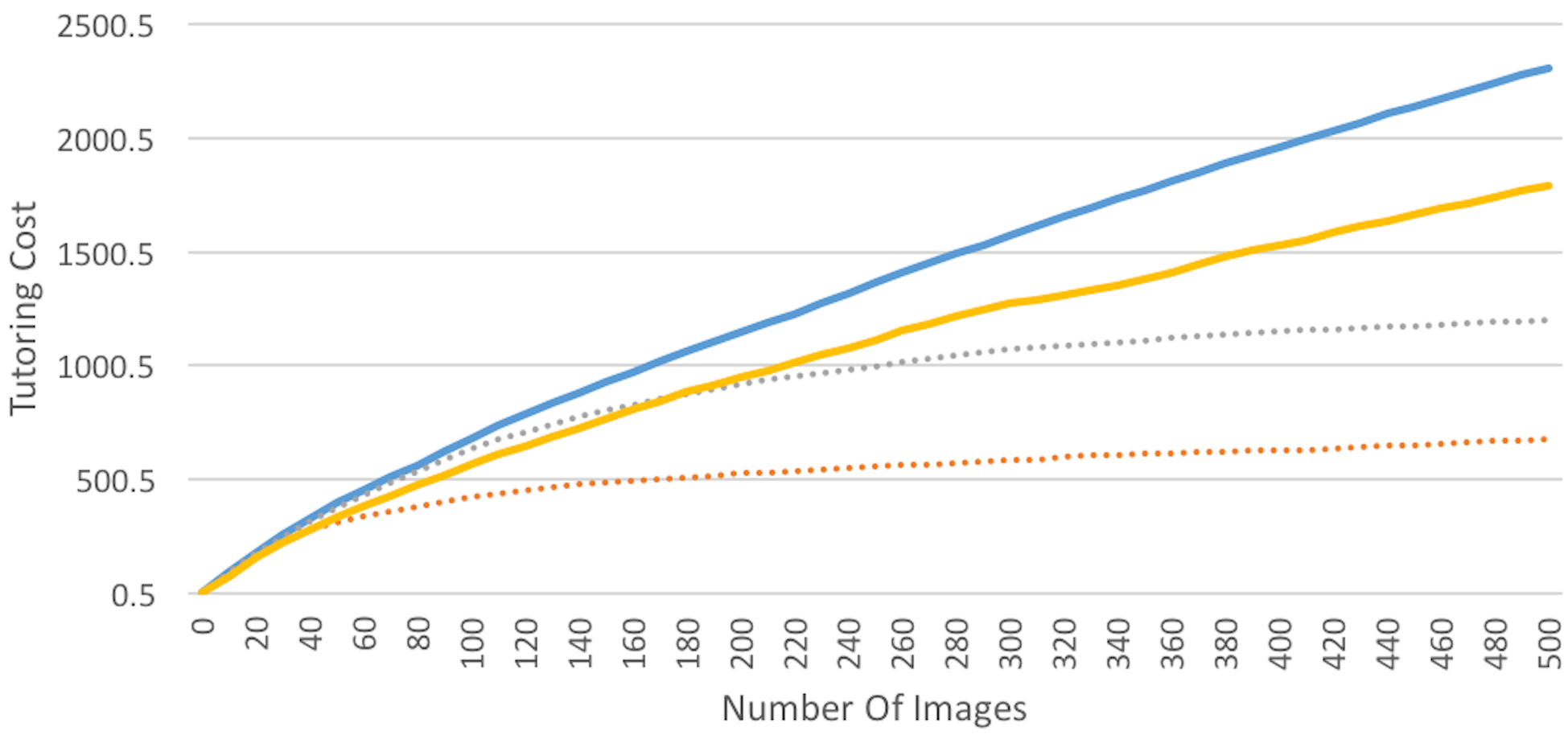}}\hfill
  
\centering  
\subfloat[Overall Performance\label{fig:ratio_expt}]
 {\includegraphics[width=.95\linewidth]{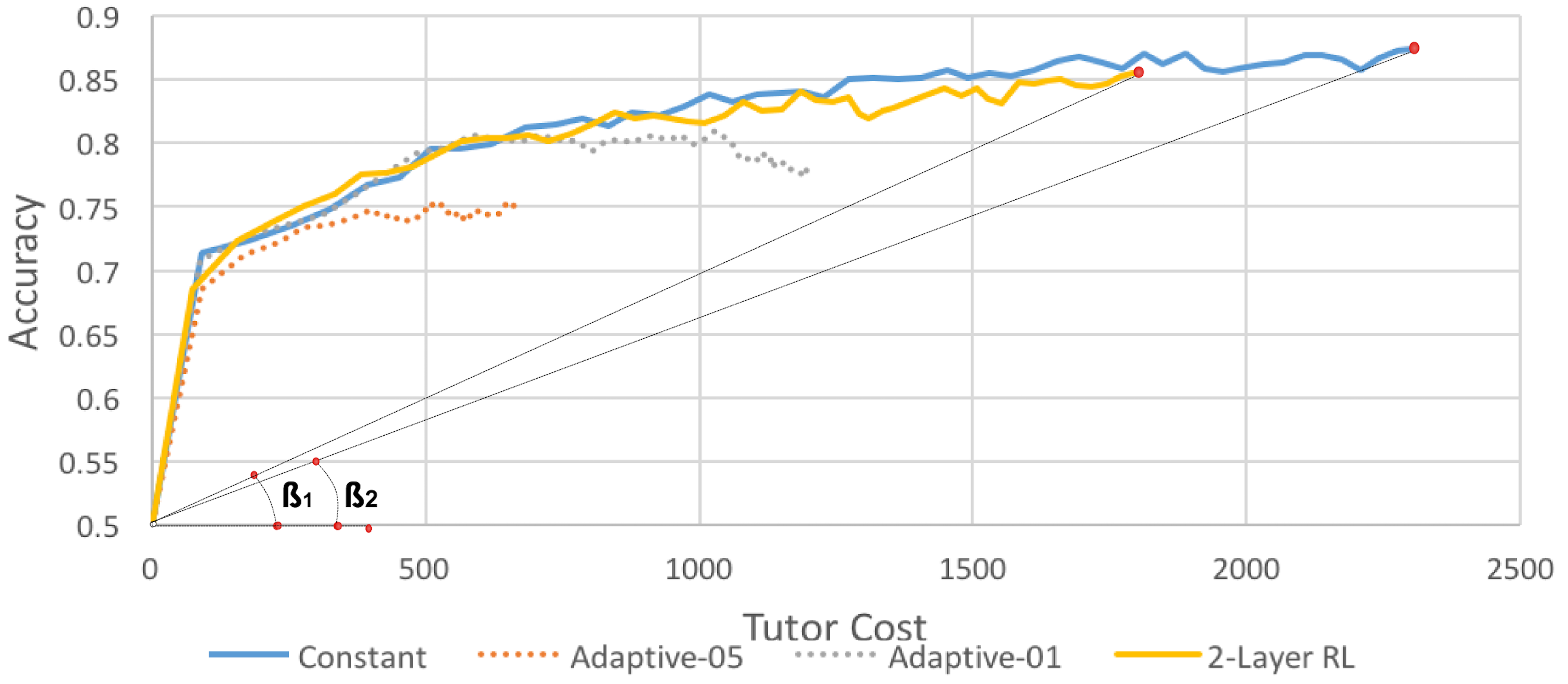}}
\caption{Evolution of Learning Performance \label{fig:results}}
\end{figure*}

Fig.\ \ref{fig:acc_expt} and \ref{fig:cost_expt} plot the progression of average Accuracy and (cumulative) Tutoring Cost for each of the 4 learning agents in our experiment, as the system interacts over time with the tutor about each of the 500 training instances. 


As noted in passing, the vertical axes in these graphs are based on averages across the 20 folds - recall that for Accuracy the system was tested, in each fold, at every learning step, i.e.\ after every 10 training instances.

Fig.\ \ref{fig:ratio_expt}, on the other hand, plots Accuracy against Tutoring Cost directly. Note that it is to be expected that the curves should not terminate in the same place on the x-axis since the different conditions incur different total costs for the tutor across the 500 training instances. The gradient of this curve corresponds to \textit{increase in Accuracy per unit of the Tutoring Cost}. It is the gradient of the line drawn from the beginning to the end of each curve ($tan(\beta)$ on Fig. \ref{fig:ratio_expt}) that constitutes  our main evaluation measure of the system's overall performance in each condition, and it is this measure for which we report statistical significance results:
there are significant differences in accuracy between the RL-based policy and two rule-based policies with the hand-crafted threshold ($p < 0.01$ for both). The RL-based policy shows  significantly less tutoring cost than the rule-based system with a constant threshold ($p < 0.01$). The mean gradient of the yellow, RL curve is actually slightly higher than the constant-threshold policy blue curve - discussed below.



\vspace{-0.15cm}

\begin{table}[!t]
\centering
\begin{tabular}{l}
\hline
\multicolumn{1}{c}{\textbf{Dialogue Example (a)}} \\
\hline
\multicolumn{1}{c}{\begin{tabular}[c]{@{}l@{}}T: what is this object called?\\ L: a red square?\\ T: the shape is correct, but the colour is wrong.\\ L: so what colour is this?\\ T: green.\\ L: okay, get it.\end{tabular}} \\ \hline \hline

\multicolumn{1}{c}{\textbf{Dialogue Example (b)}} \\
\hline
\multicolumn{1}{c}{\begin{tabular}[c]{@{}l@{}}L: blue? \\T: yes, blue is for the colour. and shape? \qquad \qquad\\L: sorry, i don't know the shape.\\T: the shape is circle. \\ L: okay, got it.\end{tabular}} \\ \hline
\end{tabular}
\vspace{-0.1cm}
\caption{User Simulation Examples for \textit{(a) Tutor takes the initiative (b) Learner takes the initiative} \label{tab:example_dialogue}}
\end{table}
\vspace{-0.2cm}

\subsection{Discussion}
\vspace{-0.1cm}
\paragraph{Accuracy} As can be seen in Fig.\ \ref{fig:acc_expt}, the rule-based system with a constant threshold ($0.95$) shows the fastest increase in accuracy and finally reaches around 0.87 at the end of the learning process (i.e.\ after seeing 500 instances) -- the blue curve. Both systems with a hand-crafted adaptive threshold, with an incremental decrease of 0.01 (grey curve) and 0.05 (orange curve), have shown an unexpected trend in accuracy across 500 instances, where the orange curve flattens out at about 0.76 after seeing only 50 instances, and the grey curve shows a good increase in the beginning but later drops down to about 0.77 after 150 instances. This is because   the thresholds were decreased too fast,  so that the agent cannot hear enough feedback (i.e.\ corrective attribute labels) from tutors to improve its predictions. In contrast to this, the optimised RL-based agent achieves much better accuracy (i.e.\ about 0.85) by the end of the experiment. 

\vspace{-0.2cm}
\paragraph{Tutoring Cost} As mentioned above, there is a form of \emph{active learning} taking place in the experiment: the agent can only hear feedback from the tutor if it is not confident enough about its own predictions. This also explains the slight decrease in the gradients of the curves (i.e.\ the cumulative cost for the tutor) (see Fig. \ref{fig:cost_expt}) as the agent is exposed to more and more training instances: its subjective confidence about its own predictions increases over time, and thus there is progressively less need for tutoring. In detail, the tutoring cost progresses much more slowly while the system was applying a hand-crafted adaptive threshold (i.e.\ incrementally decreases by either 0.01 or 0.05 after each bin). This is still because there were not interactions taking place at all once the threshold is lower than a certain value (for instance, 0.65), where the agent might be highly confident on all its predictions. In contrast, the RL-based agent shows a faster progress in the cumulative tutoring cost, but achieves higher accuracy. 

\vspace{-0.1cm}
\paragraph{Overall Performance} Here, we only compare the gradients of the curves between the optimised learning agent (yellow curve) and the rule-based system with a constant threshold (blue curve) in Fig. \ref{fig:ratio_expt}, because others with the incremental decreased threshold cannot achieve an acceptable learning performance. The agent with an adaptive threshold (yellow) achieves slightly better overall gradient ($tan(\beta_1)$) than the rule-based system ($tan(\beta_2)$), it achieves a comparable accuracy and does it faster. We therefore conclude that the optimised learning agent, which finds a better trade-off between the learning accuracy and the tutoring cost, is more desirable. 

\section{Conclusion \& Future Work}
We have introduced a multi-modal learning agent that can incrementally learn grounded word meanings through interaction with human tutors over time, and deploys an \emph{adaptive} dialogue policy (optimised using Reinforcement Learning). We applied a human-human dialogue dataset (i.e.\ BURCHAK) to train and evaluate the optimised learning agent.  We evaluated the   system by comparing it to a rule-based system, and results show that: 1) the optimised policy has learned to coherently interact with the simulated user to learn visual attributes of an object (e.g.\ colour and shape); 2) it achieves  comparable  learning performance to a rule-based systems, but with less tutoring effort needed from humans.  

Ongoing work further applies Reinforcement Learning 
at the word level to learn a complete, incremental dialogue policy, i.e.\ which chooses system output at the lexical level \cite{Eshghi.Lemon.2014,eshghi-lemon-kalatzis:2016:corr}. 
In addition, instead of acquiring visual concepts for toy objects (i.e.\ with simple colour and shape), the system has recently been extended to interactively learn about real object classes (e.g.\ {\tt shampoo, apple}). The latest system integrates with a \textit{Self-Organizing Incremental Neural Network} and a deep \textit{Convolutional Neural Network} to learn object classes through interaction with humans incrementally, over time. 

\section*{Acknowledgements}
 This research is  supported by the EPSRC, under grant number EP/M01553X/1 (BABBLE project\footnote{\url{https://sites.google.com/site/hwinteractionlab/babble}}),
and by the European Union's Horizon 2020 research and innovation programme under grant agreement No.\ 688147 (MuMMER project\footnote{\url{http://mummer-project.eu/}}).

\bibliographystyle{acl_natbib}

\bibliography{bib/babble,bib/all,bib/grounding,bib/thesis_all,bib/ownpaper,bib/additional}

\appendix

\end{document}